\documentclass[11pt,a4paper,table]{article}
\usepackage[hyperref]{emnlp2020}
\usepackage{times}
\usepackage{latexsym}
\usepackage{booktabs}
\usepackage{graphicx}
\usepackage{amsmath}
\usepackage{multirow}
\usepackage{pgfplots}
\usepackage{subcaption}
\usepackage{wrapfig}
\usepackage{hhline}
\usepackage{tabularx}
\usepackage{xcolor}
\usepackage{bm}

\newcommand{\ra}[1]{\renewcommand{\arraystretch}{#1}}
\newcommand{\cmmnt}[1]{\ignorespaces}

\usepackage{microtype}

\aclfinalcopy 
\def\aclpaperid{910} 

\definecolor{clr1}{RGB}{51,34,136}
\definecolor{clr2}{RGB}{17,119,51}
\definecolor{clr4}{RGB}{136,204,238}
\definecolor{clr5}{RGB}{221,204,119}
\definecolor{clr6}{RGB}{204,102,119}
\definecolor{clr8}{RGB}{136,34,85}

\title{TeaForN: Teacher-Forcing with N-grams}

\author{
  Sebastian Goodman \\
  Google Research \\
  Venice, CA 90291 \\
  \texttt{seabass@google.com} \\\And
  Nan Ding \\
  Google Research \\
  Venice, CA 90291 \\
  \texttt{dingnan@google.com} \\\And
  Radu Soricut \\
  Google Research \\
  Venice, CA 90291 \\
  \texttt{rsoricut@google.com} \\}

\date{}

\begin{document}
\maketitle
\begin{abstract}
Sequence generation models trained with teacher-forcing suffer from issues related to exposure bias and lack of differentiability across timesteps.
Our proposed method, Teacher-Forcing with N-grams (TeaForN), addresses both these problems directly, through the use of a stack of N decoders trained to decode along a secondary time axis that allows model-parameter updates based on N prediction steps.
TeaForN can be used with a wide class of decoder architectures and requires minimal modifications from a standard teacher-forcing setup.
Empirically, we show that TeaForN boosts generation quality on one Machine Translation benchmark, WMT 2014 English-French, and two News Summarization benchmarks, CNN/Dailymail and Gigaword.
\end{abstract}

\section{Introduction}

Many state-of-the-art sequence generation models are trained using a technique called teacher-forcing~\cite{Goodfellow-et-al-2016}.
Teacher-forcing is popular because it improves sample efficiency and provides training stability, but models trained with teacher-forcing are known to suffer from issues such as exposure bias~\cite{venkatraman-etal:2015,bengio-etal:2015,ding2017coldstart} and a lack of differentiability across timesteps (i.e., training updates made when decoding at time-step $t$ cannot fully propagate to time-step $t-1$). 
Previous attempts to address these issues include scheduled sampling~\cite{bengio-etal:2015}, parallel N-gram prediction~\cite{Yan2020}, and sampling from previous predictions~\cite{Zhang2019Bridging}.

Our proposed method, Teacher-Forcing with N-grams (TeaForN), imposes few requirements on the decoder architecture and does not require curriculum learning or sampling model outputs.
TeaForN fully embraces the teacher-forcing paradigm and extends it to N-grams, thereby addressing the problem at the level of teacher-forcing itself.

The advent of large-scale pretraining has pushed the state-of-the-art on Natural Language benchmarks to impressive heights, often showing gains across many tasks at once~\cite{Devlin2019,Raffel2019,Zhang2019}.
A negative consequence of this is the tendency towards large, data-hungry models, which have a negative impact on energy-consumption and accessibility~\citep{Strubell2019}, as well as higher latency and production costs.
As such, it is of increasing importance to develop techniques that counteract these tendencies. 
While TeaForN does increase training cost moderately, it can be used to drive down latency and inference cost, which dominate the overall cost of a production model.

Many sequence generation models use beam search to improve generation quality~\cite{vaswani2017attention,Raffel2019,Zhang2019,Yan2020}.
In contrast with greedy decoding, beam search keeps the $k$ most-likely candidates at each decoding timestep.
While beam search has proven to be a reliable technique for improving output quality, previous work has shown that beam search actually degrades performance for sufficiently large $k$~\cite{Koehn2017}.
In addition, the inference cost of a model increases linearly with $k$, due to the need for multiple decodings.
We conduct an analysis of the effect of beam size on models trained both with and without TeaForN.
We show that models trained with TeaForN require a smaller beam size to reach similar performance, a property that can achieve significant cost-savings.

Our experiments show that TeaForN can boost performance on both Machine Translation and News Summarization tasks, provided there is sufficient model capacity.
With TeaForN, Transformer~\textsubscript{big}~\cite{vaswani2017attention} improves by +.5 SacreBLEU~\cite{Post2018} on the WMT14 En-Fr benchmark with beam search and +.3 without.
When using TeaForN for summarization, PEGASUS~\textsubscript{large}~\cite{Zhang2019} improves by +.3 ROUGE-L on the Gigaword benchmark~\cite{Rush2015} and by +.2 on the CNN/Dailymail benchmark~\cite{Hermann2015}.
Further, PEGASUS~\textsubscript{large} trained with TeaForN matches the prior ROUGE-L scores on these benchmarks \textit{without beam search}, representing an 8x reduction in decoder inference cost.

\section{Related Work}
One of the standard approaches to sequence-learning training is Maximum-likelihood Estimation (MLE).
Although widely used in large array of applications, MLE estimation for sequence learning suffers from the exposure-bias problem~\cite{venkatraman-etal:2015,mixer15}.
Exposure-bias produces brittle models due to training procedures during which the models are only exposed to their training data distribution but not to their own predictions.
Possible solutions to the exposure-bias problem in neural-network settings have used ``data as demonstrator''~\cite{venkatraman-etal:2015} and ``scheduled sampling''~\cite{bengio-etal:2015} approaches.
Although improving model performance in practice, such proposals have been shown to be statistically inconsistent~\cite{huszar:2015}, and still need to perform MLE-based warm-start training, rendering such solutions unsatisfactory.
Along similar lines, the ``professor forcing''~\cite{lamb2016} method uses adversarial domain adaptation to encourage network dynamics to be the same during training and inference, though it requires sampling sequences during training.

A different approach, based on reinforcement learning methods, achieves sequence learning following a policy-gradient (PG) method~\cite{sutton-etal:1999}.
It directly attacks the exposure-bias problem by having the training models exposed exclusively to their own predictions while scoring them using reward functions.
However, this approach introduces another issue, related to the large discrepancy between the model prediction distribution and the reward function's values, which is especially acute during the early training stages when the predicted outputs are all equally bad.
As a result, this method also requires a warm-start phase in which the model distribution achieves some local maximum with respect to a reward--free objective (e.g., MLE),
followed by a model refinement phase in which reward-based PG updates are used to refine the model~\cite{mixer15,gnmt:16-short,liu-etal:2016}.
Although such combinations achieve better results in practice compared to pure likelihood-based approaches,
they are unsatisfactory from a theoretical and modeling perspective, as well as inefficient from a speed-to-convergence perspective.
A pure PG formulation that side-steps these issues is~\cite{ding2017coldstart}, which allows for both cold-start training as well as more efficient convergence properties.

The PG-based approaches have an inherent complexity that stems from the use of quirky reward functions such as ROUGE~\cite{lin2004rouge} or CIDEr~\cite{cider}, which forfeits the advantage of sample efficiency as they often cannot be efficiently computed using current accelerators like TPUs~\cite{YouZHDK19}.
MLE-based approaches appear to be favored due to efficiency properties, and the search for training methods that produce less brittle models is still on-going. 

Another closely related idea is End-to-End Backprop (E2E)~\cite{mixer15}, which has a similar goal of naturally approximating sequence level training by propagating smooth model predictions instead of groundtruth inputs.
TeaForN differs from E2E in several key ways.
First, TeaForN learns jointly from both groundtruth and model predictions as inputs throughout the entire training duration, whereas E2E requires a training schedule to transition from groundtruths to model predictions.
Second, TeaForN supports methods other than k-max for computing smooth model predictions, two of which we explore as a part of our work.
Third, we introduce the notion of a discount factor, which weights the importance of immediate predictions higher than that of future predictions.

Another such work is~\cite{Yan2020}, which proposes a modified Transformer for parallel N-gram prediction.
While their work does address the issue of strong local correlations caused by teacher-forcing, it does not address exposure bias, as it always trains on groundtruth inputs.

Also related are models such as the one proposed by~\cite{Strubell2017}, which uses a stacked of dilated convolutions to iteratively refine model predictions.
Though architecturally similar, TeaForN only uses the stack at training time and solves for a fundamentally different problem.

Our TeaForN method maintains the efficiency advantages of MLE-based approaches, while addressing both exposure bias and the issue of differentiability across timesteps.
In addition, it is general enough to be used on a wide class of autoregressive decoders, including RNN~\cite{hochreiter1997long,chung2014empirical} and Transformer~\cite{vaswani2017attention} decoders, though our experiments focus on the Transformer.

\section{TeaForN}

\begin{figure}
\centering
\includegraphics[width=.45\textwidth]{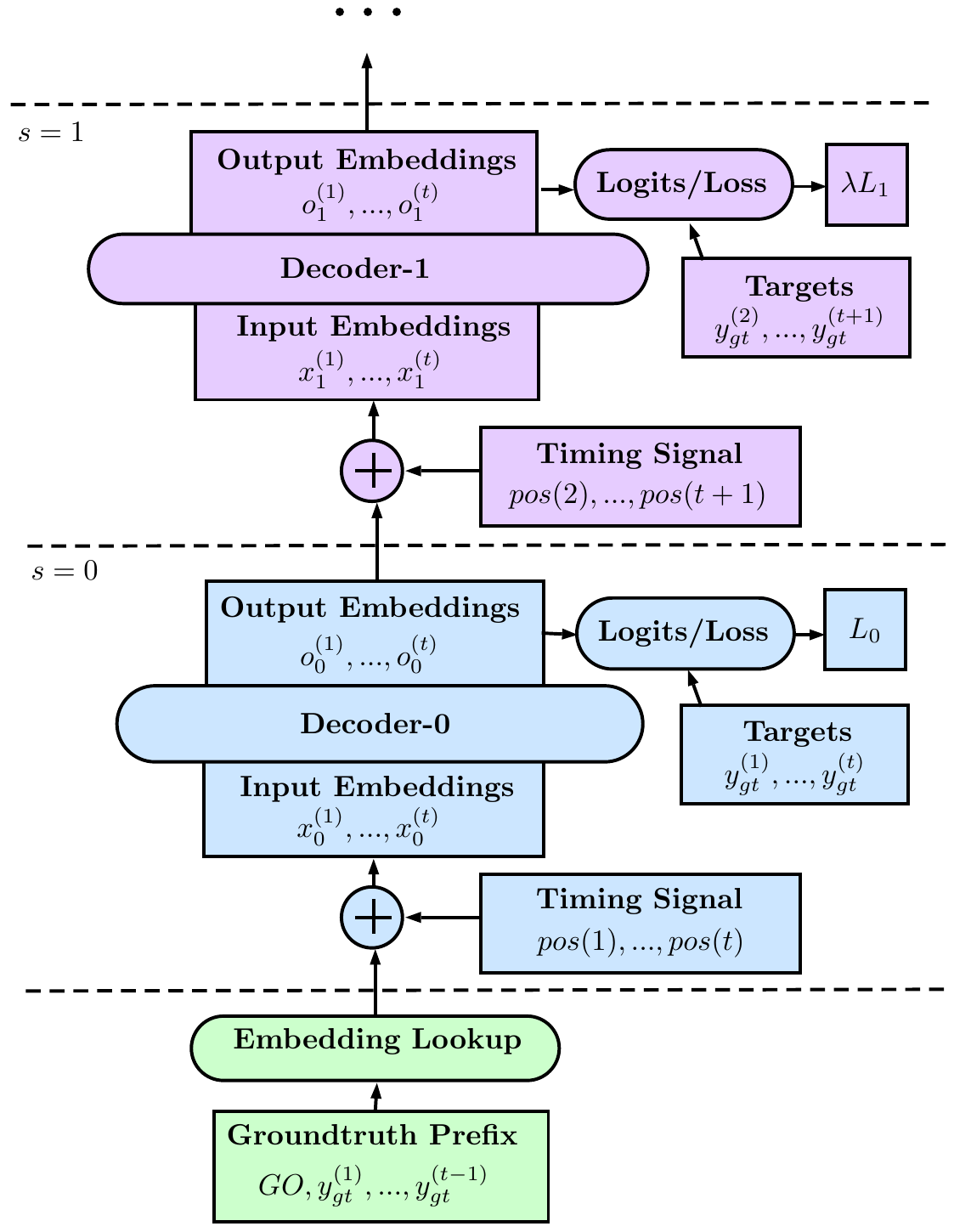}
\caption{
An illustration of TeaForN training, wherein each decoder after the first uses the outputs of the previous decoder as inputs.
Decoder weights may be shared across layers in order to address exposure bias.}
\label{fig:teafor2}
\end{figure}

Autoregressive sequence decoders are trained to minimize the negative log likelihood of the groundtruth tokens $y_{gt}^{(t)}$. During training, previous \textit{groundtruth tokens} are used as decoder inputs for predicting the next token. If we define the embedding matrix to be $E$ of size $V \times D$, where $V$ is the vocabulary size and $D$ is the embedding size, and the embedding of the groundtruth token as $x^{(t)} = E[y_{gt}^{(t-1)},:] := e_{gt}^{(t-1)}$ of size $D$, then the standard teacher-forcing loss is equal to,
\begin{align*}
L &= -\sum_{t=1}^{T}{\log(P(y_{gt}^{(t)}|y_{gt}^{(0)},...,y_{gt}^{(t-1)}))}\nonumber\\
&= -\sum_{t=1}^{T}{\log(P(y_{gt}^{(t)}|x^{(1)},...,x^{(t)}))} 
\end{align*}
where we define $y_{gt}^{(0)} = GO$ as the starting token.

The class probability distribution $P$ is typically modeled as softmax-normalized logits, which are a linear projection of decoder output $o^{(t)}$ of size $D$ onto the class embeddings using output projection matrix $W$ of size $V \times D$:
\begin{align*}
P(y^{(t)}|x^{(1)},...,x^{(t)}) = softmax(Wo^{(t)}).
\end{align*}
To reduce the model parameter size, it is standard to share the parameters of the output projection matrix and the embedding matrix, such that $E = W$.

During inference, groundtruth tokens become unavailable. Therefore, previous tokens from the \textit{model predictions} are used as decoder inputs for decoding the next token. The discrepancy between training time and inference time input distributions causes models to suffer from exposure bias, meaning that they do not learn to correct for past decoding errors~\cite{bengio-etal:2015}.

TeaForN addresses exposure bias by learning jointly how to predict from both groundtruth and past model predictions as inputs.
TeaForN setups consist of a stack of $N$ decoders, as illustrated in Figure~\ref{fig:teafor2}.
At position $t$, the first decoder (Decoder-0) takes input from the embedding of the previous groundtruth token $x_0^{(t)} = e_{gt}^{(t-1)}$ and learns to predict the target token $y_{gt}^{(t)}$, same as in teacher-forcing.
The next decoder (Decoder-1) takes input from the output $x_1^{(t)} = o_0^{(t)}$ of the first decoder and learns to predict the next target token $y_{gt}^{(t+1)}$.

More formally, let us use subscript $s \in [ 0,N )$ to denote the offset within the decoder stack.
We define the input to decoder $s$ at time $t$ as:
\begin{equation*}
\label{inputs}
x_{s}^{(t)} = pos(t+s) + \begin{cases}
e_{gt}^{(t-1)} &s=0\\
o_{s-1}^{(t)}      &s>0\\
\end{cases}
\end{equation*}
where $pos(t+s)$ is a timing signal that is added to the inputs for models such as the Transformer~\cite{vaswani2017attention}.
This term may be omitted for models that do not expect it.

The training loss of Decoder-$s$ at time $t$ is the negative log likelihood of the $(t+s)^{th}$ element in the groundtruth sequence:
\begin{equation*}
\label{loss}
L_{s} =
-\sum_{t=1}^{T}{\log(P(y_{gt}^{(t+s)}|x_{s}^{(1)},...,x_{s}^{(t)};\theta_s))}
\end{equation*}
and the total TeaForN training loss is the sum of decoder losses 
\begin{align*}
L = \sum_{s=0}^{N-1} \lambda^{s-1} L_s
\end{align*}
where $\lambda \in (0,1]$ is a discount factor needed to weigh the risk of harming next-word accuracy against the benefits of TeaForN.
During inference, TeaForN uses only the first decoder (Decoder-0) in the stack; the rest are discarded.

The intuition behind TeaForN is as follows.
Under standard teacher-forcing, the decoder output $o^{(t)}$ only learns to predict the groundtruth label $y_{gt}^{(t)}$, while outputs that favor other classes are considered equally bad, and will be penalized by the loss.
This is not reasonable because classes carrying similar meanings to the groundtruth label do not change the meaning of the sequence significantly, and may still lead to the correct prediction for the next label. 
Under TeaForN, the decoder output $o^{(t)}$ is also used as the input of a secondary decoder for decoding the next position. 
Therefore, all outputs that result in predicting the next groundtruth label $y_{gt}^{(t+1)}$ will have lower loss and therefore be differentiated from other outputs. 

In our experiments, we allow the decoder parameters to be either shared ($\theta_0 = \theta_{s}, \forall s$) or unshared.
In a shared-weight configuration, the model learns to predict the next groundtruth label from the class that the same model predicted in the previous position.
This is similar to the inference time condition, so we expect shared-weight TeaForN to address exposure-bias better than unshared-weight TeaForN.
Shared-weight configurations also have performance advantages such as lower memory consumption and faster training.

Since TeaForN solves for a more difficult problem than teacher-forcing, we expect it to work better for models with higher capacity.
We later show evidence of this by comparing results for two model sizes on Machine Translation.

It is straightforward to show that TeaFor1 (N=1) and teacher-forcing are equivalent, as the inputs to the first TeaForN decoder are groundtruth sequence embeddings and $\lambda^0=1$.
Thus, TeaForN is a natural extension of teacher-forcing to N-grams.

\subsection{Embedded Top-k Stacked Decoder Input}
Previously, our TeaForN model directly used the decoder output of the $(s-1)$-th stack as the input of the decoder of the $s$-th stack: 
\begin{align}
    x_s^{(t)} = o_{s-1}^{(t)}. \label{eq:teaforn}
\end{align}
This is an approximation to the inference-time decoder input, which (for greedy decoding) is
\begin{align}
    x_s^{(t)} = E[argmax(W o_{s-1}^{(t)}), :], \label{eq:e2e_k1}
\end{align}
where $argmax(x)$ returns the index of the $V$-dim vector with the maximum value.

Inspired by the End-to-End Backprop (E2E)~\cite{mixer15}, we also consider the following alternative decoder input,
\begin{align}
    x_s^{(t)} = E^{\top} softmax(top\_k(W o_{s-1}^{(t)})), \label{eq:e2e_k}
\end{align}
where $top\_k$ is a function which keeps the top-k values of the vector, and masks out the others.

It is easy to verify, when $k=1$, Eq.~\eqref{eq:e2e_k} reduces to Eq.~\eqref{eq:e2e_k1}; when $k=V$, 
\begin{align*}
    x_s^{(t)} = E^{\top} softmax(W o_{s-1}^{(t)}).
\end{align*}

Compared to Eq.~\eqref{eq:teaforn}, Eq.~\eqref{eq:e2e_k} is more computationally expensive, as it involves additional embedding matrix multiplications and/or a top-k sorting. Furthermore, we would like to emphasize a critical difference between the TeaForN and E2E~\cite{mixer15}. In TeaForN, the 0-th stack of every position is always clamped to the groundtruth input, while for E2E the groundtruth is completely thrown away after warm-up training. The groundtruth clamping allows the TeaForN to avoid the warm-up training which is necessary for E2E. 

\begin{table*}[!htbp]
\begin{centering}
\begin{tabular}{ @{} rccccccc @{}}\toprule
\ra{1.3}
\multirow{2}{*}{} & \multirow{2}{*}{$\theta_{shared}$} & \phantom{abc} & \multicolumn{2}{c}{Greedy decoding} & \phantom{abc} & \multicolumn{2}{c}{Beam search@k=4} \\
\cmidrule{4-5}
\cmidrule{7-8}
&&& Teacher-forcing & TeaFor2 && Teacher-forcing & TeaFor2 \\
\midrule
\multirow{2}{*}{En-De} 
& \multirow{1}{*}{N}
&& \multirow{2}{*}{$26.96 \pm .04$} & $27.02 \pm .06$
&& \multirow{2}{*}{$27.96 \pm .09$} & $27.88 \pm .05$ \\
& \multirow{1}{*}{Y}
&&& $\bm{27.16} \pm .02$
&&& $27.90 \pm .03$ \\
\hline
\multirow{2}{*}{En-Fr} 
& \multirow{1}{*}{N}
&& \multirow{2}{*}{$40.20 \pm .04$} & $\bm{40.32} \pm .08$
&& \multirow{2}{*}{$40.86 \pm .08$} & $40.88 \pm .10$ \\
& \multirow{1}{*}{Y}
&&& $\bm{40.32} \pm .04$
&&& $40.84 \pm .07$ \\
\bottomrule
\end{tabular}
\caption{
A comparison of models on WMT14 language pairs En-De and En-Fr using Transformer\textsubscript{base}.
We report mean and Standard Error of SacreBLEU scores over five independent training runs.
$\theta_{shared}$ refers to whether the free parameters of the decoder are shared across decoder instances (Y) or kept separate (N).
The discount factor is $\lambda=.5$ for TeaFor2 models.
}
\label{baseresults}
\end{centering}
\end{table*}

\begin{table}[!b]
\begin{centering}
\begin{tabular}{ @{} rrr @{} } \toprule
\ra{1.3}
  & Transformer\textsubscript{base} & Transformer\textsubscript{big} \\
\midrule
$P_{drop}$  & .1 & .3 \\
$d_{model}$ & 512 & 1024 \\
$d_{ff}$ & 2048 & 4096 \\
$h$ & 8 & 16 \\
 \bottomrule
\end{tabular}
\caption{
A summary of differences between model variants Transformer\textsubscript{base} and Transformer\textsubscript{big}.
$P_{drop}$ refers to dropout probability, $d_{model}$ refers to class embedding size and hidden size, $d_{ff}$ refers to the size of feedforward layers, and $h$ refers to the number of self-attention heads~\cite{vaswani2017attention}.
}
\label{transformersizes}
\end{centering}
\end{table}

\section{Experimental Results}

Our empirical study of TeaForN is comprised of two sections.
First, we present experiments on Machine Translation using the well-known Transformer model~\cite{vaswani2017attention}.
Second, we show results for News Summarization, for which we use PEGASUS~\cite{Zhang2019}, a state-of-the-art pretrained text summarization model.

We perform minimal hyperparameter tuning over the course of these experiments.
This can be partly credited to the underlying models being well-tuned already, but also to TeaForN, which works out-of-the-box without much hyperparameter tuning.
One exception is the tuning of the number of training steps, as we found that the number of steps used by previous settings is sometimes insufficient.

\subsection{Machine Translation}
In this section, we study the effects of applying TeaForN to  a well-known Transformer-based Machine Translation model.
We present results for two size variants of the model, Transformer\textsubscript{base} and Transformer\textsubscript{big}~\cite{vaswani2017attention}.
The differences are summarized in Table~\ref{transformersizes}.

We use the same WMT14 language-pair benchmarks originally reported in the Transformer paper:
\begin{itemize}
    \item \textbf{English-German (En-De)}, with 4.5M sentence pairs for training and 2,737 for testing.
    \item \textbf{English-French (En-Fr)}, with 36M sentence pairs for training and 3,003 for testing.
\end{itemize}

We use \textit{SacreBLEU}~\cite{Post2018} with case-sensitive tokenization to score translations. 
We report SacreBLEU scores for beam search widths $k \in [1,8]$ to show the interaction between TeaForN learning and beam search.

\subsubsection{Transformer\textsubscript{base}}
Using Transformer\textsubscript{base} as our underlying model, we measure the impact of TeaForN on the Machine Translation task.
We test both shared- and unshared-weight configurations, with $N=2$ (i.e. "TeaFor2") and $\lambda=.5$.
We expect weight-shared configurations to be more effective, as a more direct means of addressing exposure bias in the decoder.

All models are trained for 1M steps, and we observe no signs of overfitting.
For model selection, we average the last five checkpoints, as originally done for the Transformer~\cite{vaswani2017attention}.
We report mean and standard-error variation of SacreBLEU scores over five runs.

Table~\ref{baseresults} shows that TeaFor2 improves the quality of greedy decoding on both language pairs.
TeaFor2 raises SacreBLEU scores by +.20 on En-De (27.16 vs 26.96) and +.12 on En-Fr (40.32 vs 40.20).
Shared-weight TeaFor2 boosts performance  on the En-De benchmark by +.14 SacreBLEU (27.16 vs 27.02).
The small size of the En-De training set (relative to En-Fr), means that the En-De model has additional  capacity for learning the TeaFor2 task.
This supports our case that TeaForN with weight-sharing improves model performance, but only if there is sufficient model capacity.
Table~\ref{baseresults} also shows that TeaFor2, with or without weight-sharing for En-Fr, outperforms standard teacher-forcing by the same amount, +.12 SacreBLEU (40.32 vs 40.20).
We credit the increase in performance to TeaForN's ability to make predictions that lead to better predictions in the subsequent sequence positions.

Beam search results in Table~\ref{baseresults} show that the gains of TeaFor2 are negated by beam search with $k=4$.
Because of the small capacity of the Transformer\textsubscript{base} model, the benefits of TeaFor2 are minimal and only reflected in the result from greedy decoding.
In the following experiment, we show that higher-capacity models benefit more from TeaForN when using beam search.

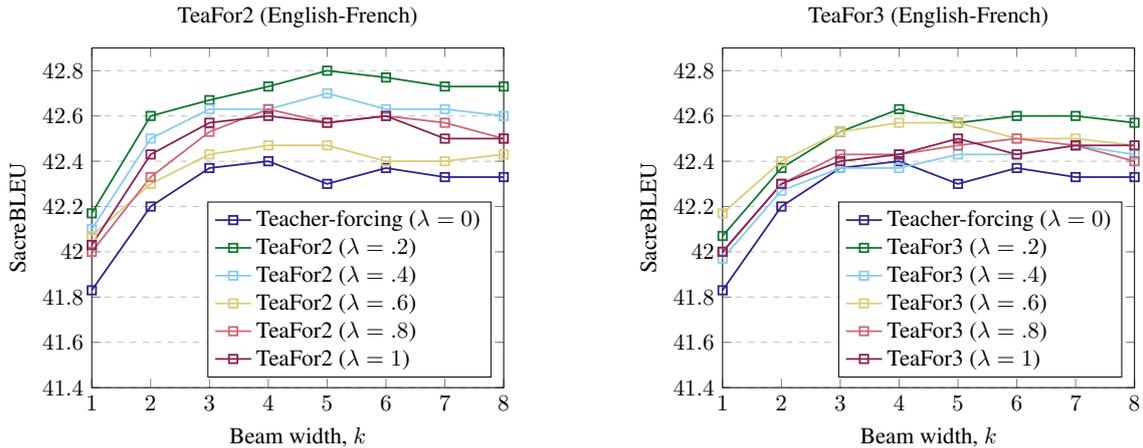
\begin{figure*}[!htpb]
    \begin{subfigure}{\columnwidth}
        \centering
        \resizebox{.9\columnwidth}{!}{
            \begin{tikzpicture}
            \begin{axis}[
                title={TeaFor2 (English-French)},
                xlabel={Beam width, $k$},
                ylabel={SacreBLEU},
                xmin=1, xmax=8,
                ymin=41.4, ymax=42.9,
                xtick distance=1,
                ytick distance=.2,
                legend pos=south east,
                ymajorgrids=true,
                grid style=dashed,
                legend cell align=left,
            ]
            \addplot[color=clr1, line width=.3mm, mark=square] coordinates {
            (1,	41.83)
            (2,	42.20)
            (3,	42.37)
            (4,	42.40)
            (5,	42.30)
            (6,	42.37)
            (7,	42.33)
            (8,	42.33)
            };
            \addlegendentry{Teacher-forcing ($\lambda=0$)}
            \addplot[color=clr2, line width=.3mm, mark=square] coordinates {
            (1, 42.17)
            (2,	42.60)
            (3,	42.67)
            (4,	42.73)
            (5,	42.80)
            (6,	42.77)
            (7,	42.73)
            (8,	42.73)
            };
            \addlegendentry{TeaFor2 ($\lambda=.2$)}
            \addplot[color=clr4, line width=.3mm, mark=square] coordinates {
            (1, 42.10)
            (2,	42.50)
            (3,	42.63)
            (4,	42.63)
            (5,	42.70)
            (6,	42.63)
            (7,	42.63)
            (8,	42.60)
            };
            \addlegendentry{TeaFor2 ($\lambda=.4$)}
            \addplot[color=clr5, line width=.3mm, mark=square] coordinates {
            (1, 42.07)
            (2,	42.30)
            (3,	42.43)
            (4,	42.47)
            (5,	42.47)
            (6,	42.40)
            (7,	42.40)
            (8,	42.43)
            };
            \addlegendentry{TeaFor2 ($\lambda=.6$)}
            \addplot[color=clr6, line width=.3mm, mark=square] coordinates {
            (1, 42.0)
            (2,	42.33)
            (3,	42.53)
            (4,	42.63)
            (5,	42.57)
            (6,	42.60)
            (7,	42.57)
            (8,	42.50)
            };
            \addlegendentry{TeaFor2 ($\lambda=.8$)}
            \addplot[color=clr8, line width=.3mm, mark=square] coordinates {
            (1, 42.03)
            (2,	42.43)
            (3,	42.57)
            (4,	42.60)
            (5,	42.57)
            (6,	42.60)
            (7,	42.50)
            (8,	42.50)
            };
            \addlegendentry{TeaFor2 ($\lambda=1$)}
            \end{axis}
            \end{tikzpicture}
        }
    \end{subfigure}%
    \hfill%
    \begin{subfigure}{\columnwidth}
        \centering
        \resizebox{.9\columnwidth}{!}{
            \begin{tikzpicture}
            \begin{axis}[
                title={TeaFor3 (English-French)},
                xlabel={Beam width, $k$},
                ylabel={SacreBLEU},
                xmin=1, xmax=8,
                ymin=41.4, ymax=42.9,
                xtick distance=1,
                ytick distance=.2,
                legend pos=south east,
                ymajorgrids=true,
                grid style=dashed,
                legend cell align=left,
            ]
            \addplot[color=clr1, line width=.3mm, mark=square] coordinates {
            (1,	41.83)
            (2,	42.20)
            (3,	42.37)
            (4,	42.40)
            (5,	42.30)
            (6,	42.37)
            (7,	42.33)
            (8,	42.33)
            };
            \addlegendentry{Teacher-forcing ($\lambda=0$)}
            \addplot[color=clr2, line width=.3mm, mark=square] coordinates {
            (1, 42.07)
            (2,	42.37)
            (3,	42.53)
            (4,	42.63)
            (5,	42.57)
            (6,	42.60)
            (7,	42.60)
            (8,	42.57)
            };
            \addlegendentry{TeaFor3 ($\lambda=.2$)}
            \addplot[color=clr4, line width=.3mm, mark=square] coordinates {
            (1, 41.97)
            (2,	42.27)
            (3,	42.37)
            (4,	42.37)
            (5,	42.43)
            (6,	42.43)
            (7,	42.47)
            (8,	42.43)
            };
            \addlegendentry{TeaFor3 ($\lambda=.4$)}
            \addplot[color=clr5, line width=.3mm, mark=square] coordinates {
            (1, 42.17)
            (2,	42.40)
            (3,	42.53)
            (4,	42.57)
            (5,	42.57)
            (6,	42.50)
            (7,	42.50)
            (8,	42.47)
            };
            \addlegendentry{TeaFor3 ($\lambda=.6$)}
            \addplot[color=clr6, line width=.3mm, mark=square] coordinates {
            (1, 42.00)
            (2,	42.30)
            (3,	42.43)
            (4,	42.43)
            (5,	42.47)
            (6,	42.50)
            (7,	42.47)
            (8,	42.40)
            };
            \addlegendentry{TeaFor3 ($\lambda=.8$)}
            \addplot[color=clr8, line width=.3mm, mark=square] coordinates {
            (1, 42.00)
            (2,	42.30)
            (3,	42.40)
            (4,	42.43)
            (5,	42.50)
            (6,	42.43)
            (7,	42.47)
            (8,	42.47)
            };
            \addlegendentry{TeaFor3 ($\lambda=1$)}
            \end{axis}
            \end{tikzpicture}
        }
    \end{subfigure}\vspace{8mm}
    \caption{Beam width vs. SacreBLEU on the WMT 2014 English-French benchmark, for discount factors $\lambda \in \{0,.2,.4,.6,.8,1\}$.
    Reported scores are averages over n=3 independent training runs, with error bars omitted for readability.
    See Appendix Tables~\ref{teafor2enfr14} and~\ref{teafor3enfr14} for results with standard error measurements.}
    \label{bigplotsenfr}
\end{figure*}

\begin{table}[!b]
\begin{centering}
\begin{tabular}{ @{} r rr @{} } \toprule
\ra{1.3}
& En-De & En-Fr\\
\midrule
~\cite{Ott2018} & $28.6$ & $41.4$ \\
~\cite{So2019} & $29.2$ & - \\
\hline
Transformer\textsubscript{big} & $29.20 \pm 0.12$ & $42.33 \pm 0.07$ \\
TeaFor2 & $29.30 \pm 0.05$ & $\bm{42.73} \pm 0.05$ \\
TeaFor3 & $29.23 \pm 0.05$ & $42.43 \pm 0.03$ \\
 \bottomrule
\end{tabular}
\caption{
A comparison of SacreBLEU scores of Machine Translation models on WMT14 En-De and En-Fr benchmarks.
Results for our models are shown below the horizontal line.
We report mean and standard error over n=3 independent training runs.
We set beam width $k=8$ for all models; we tune $\lambda$ against the validation set (selected values of $\lambda$ are .4, .2, .4, and .4 left-to-right and top-to-bottom).
See Appendix for test and validation scores with standard error measurements.
}
\label{translateleaderboard}
\end{centering}
\end{table}

\subsubsection{Transformer\textsubscript{big}}
Using Transformer\textsubscript{big}, we now compare standard teacher-forcing against TeaForN (N=2,3).

We test against the same WMT14 language pairs as the previous experiments.
We train En-Fr models for 1M steps and En-De models for 500k steps.
Beyond 500k training steps, we observe that En-De models overfit the training data (see Table~\ref{samecompute}).
This is likely due to a combination of larger model capacity in Transformer\textsubscript{big} (Table~\ref{transformersizes}) and smaller training set for En-De.
For model selection, we average the last twenty checkpoints, as was done for the Transformer~\cite{vaswani2017attention}.

We use weight-sharing for all TeaForN setups in this section.
Transformer~\textsubscript{big} has more capacity than Transformer~\textsubscript{base}, so it is expected to perform better in a shared-weight configuration.

Figure~\ref{bigplotsenfr} shows that TeaForN outperforms standard-teacher forcing on the En-Fr benchmark, across all beam widths up to $k=8$ and all discount factors $\lambda \in \{.2,.4,.6,.8,1\}$.
With beam 2, TeaFor2 achieves a higher score on En-Fr than teacher-forcing achieves with any beam size up to 8 (42.6 vs 42.4) and significantly outperforms it with beam size 5 (42.8 vs 42.4).
TeaFor3 performs as well as Teacher-forcing but worse than TeaFor2 on the En-Fr benchmark, for nearly every discount factor tested.
This shows that TeaForN can be used to train models with higher quality for any given beam size or, alternatively, train models of similar quality but lower inference cost (i.e., faster).

In contrast with the results for lower-capacity models, Fig.~\ref{bigplotsenfr} shows that beam search does not erase gains due to TeaForN training.
The Teacher-forcing setup gains +.6 SacreBLEU from beam search (42.4 vs 41.8) compared to +.6 for TeaFor2 (42.8 vs 42.2), in spite of a +.4 SacreBLEU higher baseline.
Provided sufficient model capacity, TeaForN is seen to improve the quality of the underlying model, so that greedy decoding is more effective, but not at the expense of beam search.

Intuitively, discount factors that are too high may interfere with prediction quality, as they decrease the relative importance of next word prediction.
We see this on the English-German benchmark, shown in Fig.~\ref{bigplotsende}, where the highest discount factor tested ($\lambda=1$) significantly reduces greedy performance (27.9/27.8 from 28.1) and peak performance (28.9/28.8 vs 29.2).
In all of our Transformer\textsubscript{big} experiments, the best performing discount factor is either .2 or .4, which are the lowest values tested.

Table~\ref{translateleaderboard} shows our results compared to current state-of-the-art Machine Translation models.
To allow for a fair comparison, we select our discount factor $\lambda \in \{.2,.4,.6,.8,1\}$ to maximize performance against a development set, WMT12.
On the En-De benchmark, TeaForN setups perform similarly to Teacher-forcing, at 29.0 SacreBLEU.
On En-Fr, TeaFor2 outperforms Teacher-forcing significantly, by +.4 SacreBLEU (42.7 vs 42.3).

\begin{table}[!b]
\begin{centering}
\begin{tabular}{ @{} llrr @{} } \toprule
\ra{1.3}
& & 1x iterations & 1.5x iterations \\
\midrule
\multirow{2}{*}{En-De} & greedy & $28.10 \pm .12$ & $27.97 \pm .20$ \\
                       & beam   & $29.30 \pm .20$ & $29.20 \pm .16$ \\
\midrule
\multirow{2}{*}{En-Fr} & greedy & $41.83 \pm .08$ & $41.90 \pm .12$ \\
                       & beam   & $42.40 \pm .04$ & $42.40 \pm .12$ \\
 \bottomrule
\end{tabular}
\caption{
SacreBLEU scores of Transformer\textsubscript{big} on WMT14 En-De and En-Fr benchmarks.
We report mean and standard error over n=3 independent training runs.
}
\label{samecompute}
\end{centering}
\end{table}

\definecolor{lightgray}{gray}{0.9}
\definecolor{correct}{rgb}{0,.7,0}
\newcounter{tblerows}
\expandafter\let\csname c@tblerows\endcsname\rownum

\begin{figure*}[!htpb]
    \begin{subfigure}{\columnwidth}
        \centering
        \resizebox{.9\columnwidth}{!}{
            \begin{tikzpicture}
            \begin{axis}[
                title={TeaFor2 (English-German)},
                xlabel={Beam width, $k$},
                ylabel={SacreBLEU},
                xmin=1, xmax=8,
                ymin=27.6, ymax=29.4,
                xtick distance=1,
                ytick distance=.2,
                legend pos=south east,
                ymajorgrids=true,
                grid style=dashed,
                legend cell align=left,
            ]
            \addplot[color=clr1, line width=.3mm, mark=square] coordinates {
            (1,	28.08)
            (2,	28.63)
            (3,	28.95)
            (4,	29.08)
            (5,	29.18)
            (6,	29.25)
            (7,	29.20)
            (8,	29.18)
            };
            \addlegendentry{Teacher-forcing ($\lambda=0$)}
            \addplot[color=clr2, line width=.3mm, mark=square] coordinates {
            (1,	28.18)
            (2,	28.60)
            (3,	28.95)
            (4,	29.08)
            (5,	29.10)
            (6,	29.15)
            (7,	29.18)
            (8,	29.18)
            };
            \addlegendentry{TeaFor2 ($\lambda=.2$)}
            \addplot[color=clr4, line width=.3mm, mark=square] coordinates {
            (1,	28.18)
            (2,	28.80)
            (3,	29.08)
            (4,	29.28)
            (5,	29.30)
            (6,	29.33)
            (7,	29.33)
            (8,	29.35)
            };
            \addlegendentry{TeaFor2 ($\lambda=.4$)}
            \addplot[color=clr5, line width=.3mm, mark=square] coordinates {
            (1,	28.25)
            (2,	28.83)
            (3,	29.00)
            (4,	29.08)
            (5,	29.13)
            (6,	29.18)
            (7,	29.18)
            (8,	29.13)
            };
            \addlegendentry{TeaFor2 ($\lambda=.6$)}
            \addplot[color=clr6, line width=.3mm, mark=square] coordinates {
            (1,	28.18)
            (2,	28.85)
            (3,	29.03)
            (4,	29.15)
            (5,	29.20)
            (6,	29.25)
            (7,	29.30)
            (8,	29.28)
            };
            \addlegendentry{TeaFor2 ($\lambda=.8$)}
            \addplot[color=clr8, line width=.3mm, mark=square] coordinates {
            (1,	27.90)
            (2,	28.53)
            (3,	28.78)
            (4,	28.80)
            (5,	28.88)
            (6,	28.88)
            (7,	28.93)
            (8,	28.85)
            };
            \addlegendentry{TeaFor2 ($\lambda=1$)}
            \end{axis}
            \end{tikzpicture}
        }
    \end{subfigure}%
    \hfill%
    \begin{subfigure}{\columnwidth}
        \centering
        \resizebox{.9\columnwidth}{!}{
            \begin{tikzpicture}
            \begin{axis}[
                title={TeaFor3 (English-German)},
                xlabel={Beam width, $k$},
                ylabel={SacreBLEU},
                xmin=1, xmax=8,
                ymin=27.6, ymax=29.4,
                xtick distance=1,
                ytick distance=.2,
                legend pos=south east,
                ymajorgrids=true,
                grid style=dashed,
                legend cell align=left,
            ]
            \addplot[color=clr1, line width=.3mm, mark=square] coordinates {
            (1,	28.08)
            (2,	28.63)
            (3,	28.95)
            (4,	29.08)
            (5,	29.18)
            (6,	29.25)
            (7,	29.20)
            (8,	29.18)
            };
            \addlegendentry{Teacher-forcing ($\lambda=0$)}
            \addplot[color=clr2, line width=.3mm, mark=square] coordinates {
            (1,	28.40)
            (2,	29.00)
            (3,	29.27)
            (4,	29.33)
            (5,	29.30)
            (6,	29.33)
            (7,	29.37)
            (8,	29.33)
            };
            \addlegendentry{TeaFor3 ($\lambda=.2$)}
            \addplot[color=clr4, line width=.3mm, mark=square] coordinates {
            (1,	28.07)
            (2,	28.67)
            (3,	28.87)
            (4,	28.97)
            (5,	29.13)
            (6,	29.17)
            (7,	29.20)
            (8,	29.23)
            };
            \addlegendentry{TeaFor3 ($\lambda=.4$)}
            \addplot[color=clr5, line width=.3mm, mark=square] coordinates {
            (1,	28.23)
            (2,	28.73)
            (3,	29.00)
            (4,	29.00)
            (5,	29.00)
            (6,	29.03)
            (7,	29.10)
            (8,	29.07)
            };
            \addlegendentry{TeaFor3 ($\lambda=.6$)}
            \addplot[color=clr6, line width=.3mm, mark=square] coordinates {
            (1,	28.17)
            (2,	28.60)
            (3,	28.87)
            (4,	28.93)
            (5,	28.97)
            (6,	29.00)
            (7,	29.03)
            (8,	29.00)
            };
            \addlegendentry{TeaFor3 ($\lambda=.8$)}
            \addplot[color=clr8, line width=.3mm, mark=square] coordinates {
            (1,	27.80)
            (2,	28.40)
            (3,	28.63)
            (4,	28.67)
            (5,	28.70)
            (6,	28.77)
            (7,	28.73)
            (8,	28.77)
            };
            \addlegendentry{TeaFor3 ($\lambda=1$)}
            \end{axis}
            \end{tikzpicture}
        }
    \end{subfigure}
    \caption{Beam width vs. SacreBLEU on the WMT 2014 English-German benchmark, for discount factors $\lambda \in \{0,.2,.4,.6,.8,1\}$.
    Left and right plots show TeaFor2 and TeaFor3, respectively.
    Reported scores are averages over n=3 independent training runs, with error bars omitted for readability.
    See Appendix Tables~\ref{teafor2ende14} and~\ref{teafor3ende14} for results with standard error measurements.}
    \label{bigplotsende}
    \vspace{5mm}
    \begin{subfigure}{\columnwidth}
        \centering
        \resizebox{.9\columnwidth}{!}{
            \begin{tikzpicture}
            \begin{axis}[
                title={English-German},
                xlabel={Beam width, $k$},
                ylabel={SacreBLEU},
                xmin=1, xmax=8,
                ymin=27.6, ymax=29.4,
                xtick distance=1,
                ytick distance=.2,
                legend pos=south east,
                ymajorgrids=true,
                grid style=dashed,
                legend cell align=left,
            ]
            \addplot[color=clr1, line width=.3mm, mark=square] coordinates {
            (1,	28.10)
            (2,	28.67)
            (3,	28.97)
            (4,	29.07)
            (5,	29.20)
            (6,	29.30)
            (7,	29.23)
            (8,	29.20)
            };
            \addlegendentry{Teacher-forcing}
            \addplot[color=clr2, line width=.3mm, mark=square] coordinates {
            (1,	28.17)
            (2,	28.67)
            (3,	28.97)
            (4,	29.07)
            (5,	29.13)
            (6,	29.17)
            (7,	29.20)
            (8,	29.20)
            };
            \addlegendentry{TeaFor2}
            \addplot[color=clr4, line width=.3mm, mark=square] coordinates {
            (1,	27.83)
            (2,	28.57)
            (3,	28.80)
            (4,	28.90)
            (5,	28.97)
            (6,	29.00)
            (7,	29.00)
            (8,	29.00)
            };
            \addlegendentry{TeaFor2 ($top_4$)}
            \addplot[color=clr5, line width=.3mm, mark=square] coordinates {
            (1,	28.03)
            (2,	28.73)
            (3,	29.00)
            (4,	29.13)
            (5,	29.23)
            (6,	29.30)
            (7,	29.23)
            (8,	29.30)
            };
            \addlegendentry{TeaFor2 ($top_V$)}
            \end{axis}
            \end{tikzpicture}
        }
    \end{subfigure}%
    \hfill%
    \begin{subfigure}{\columnwidth}
        \centering
        \resizebox{.9\columnwidth}{!}{
            \begin{tikzpicture}
            \begin{axis}[
                title={English-French},
                xlabel={Beam width, $k$},
                ylabel={SacreBLEU},
                xmin=1, xmax=8,
                ymin=41.7, ymax=42.9,
                xtick distance=1,
                ytick distance=.2,
                legend pos=south east,
                ymajorgrids=true,
                grid style=dashed,
                legend cell align=left,
            ]
            \addplot[color=clr1, line width=.3mm, mark=square] coordinates {
            (1,	41.83)
            (2,	42.20)
            (3, 42.37)
            (4,	42.40)
            (5,	42.30)
            (6,	42.37)
            (7,	42.33)
            (8,	42.33)
            };
            \addlegendentry{Teacher-forcing}
            \addplot[color=clr2, line width=.3mm, mark=square] coordinates {
            (1,	42.17)
            (2,	42.60)
            (3,	42.67)
            (4,	42.73)
            (5,	42.80)
            (6,	42.77)
            (7,	42.73)
            (8,	42.73)
            };
            \addlegendentry{TeaFor2}
            \addplot[color=clr4, line width=.3mm, mark=square] coordinates {
            (1,	42.23)
            (2,	42.50)
            (3,	42.73)
            (4,	42.77)
            (5,	42.77)
            (6,	42.80)
            (7,	42.73)
            (8,	42.73)
            };
            \addlegendentry{TeaFor2 ($top_4$)}
            \addplot[color=clr5, line width=.3mm, mark=square] coordinates {
            (1,	41.90)
            (2,	42.27)
            (3,	42.47)
            (4,	42.47)
            (5,	42.47)
            (6,	42.47)
            (7,	42.47)
            (8,	42.43)
            };
            \addlegendentry{TeaFor2 ($top_V$)}
            \end{axis}
            \end{tikzpicture}
        }
    \end{subfigure}
    \caption{Beam width vs. SacreBLEU on WMT 2014 benchmarks comparing approximation methods Top-K.
    Reported scores are averages over n=3 independent training runs, with error bars omitted for readability.
    See Appendix Tables~\ref{approxenfr} and~\ref{approxende} for results with standard error measurements.}
    \label{topkplot}
    \vspace{5mm}
    \begin{subfigure}{\columnwidth}
        \centering
        \resizebox{.9\columnwidth}{!}{
            \begin{tikzpicture}
            \begin{axis}[
                title={English-German},
                xlabel={Beam width, $k$},
                ylabel={SacreBLEU},
                xmin=1, xmax=8,
                ymin=27.8, ymax=29.6,
                xtick distance=1,
                ytick distance=.2,
                legend pos=south east,
                ymajorgrids=true,
                grid style=dashed,
                legend cell align=left,
            ]
            \addplot[color=clr1, line width=.3mm, mark=square] coordinates {
            (1,	28.10)
            (2,	28.67)
            (3,	28.97)
            (4,	29.07)
            (5,	29.20)
            (6,	29.30)
            (7,	29.23)
            (8,	29.20)
            };
            \addlegendentry{$P_{drop}=0$ (Teacher-forcing)}
            \addplot[color=clr2, line width=.3mm, mark=square] coordinates {
            (1,	28.60)
            (2,	29.13)
            (3,	29.23)
            (4,	29.43)
            (5,	29.47)
            (6,	29.50)
            (7,	29.47)
            (8,	29.47)
            };
            \addlegendentry{$P_{drop}=.1$}
            \addplot[color=clr4, line width=.3mm, mark=square] coordinates {
            (1,	28.37)
            (2,	28.90)
            (3,	29.13)
            (4,	29.20)
            (5,	29.33)
            (6,	29.30)
            (7,	29.30)
            (8,	29.33)
            };
            \addlegendentry{$P_{drop}=.2$}
            \addplot[color=clr5, line width=.3mm, mark=square] coordinates {
            (1,	28.20)
            (2,	28.77)
            (3,	28.93)
            (4,	29.07)
            (5,	29.13)
            (6,	29.23)
            (7,	29.20)
            (8,	29.20)
            };
            \addlegendentry{$P_{drop}=.3$}
            \end{axis}
            \end{tikzpicture}
        }
    \end{subfigure}%
    \hfill%
    \begin{subfigure}{\columnwidth}
        \centering
        \resizebox{.9\columnwidth}{!}{
            \begin{tikzpicture}
            \begin{axis}[
                title={English-French},
                xlabel={Beam width, $k$},
                ylabel={SacreBLEU},
                xmin=1, xmax=8,
                ymin=41.38, ymax=42.44,
                xtick distance=1,
                ytick distance=.2,
                legend pos=south east,
                ymajorgrids=true,
                grid style=dashed,
                legend cell align=left,
            ]
            \addplot[color=clr1, line width=.3mm, mark=square] coordinates {
            (1,	41.83)
            (2,	42.20)
            (3,	42.37)
            (4,	42.40)
            (5,	42.30)
            (6,	42.37)
            (7,	42.33)
            (8,	42.33)
            };
            \addlegendentry{$P_{drop}=0$ (Teacher-forcing)}
            \addplot[color=clr2, line width=.3mm, mark=square] coordinates {
            (1,	41.63)
            (2,	41.93)
            (3,	42.13)
            (4,	42.17)
            (5,	42.10)
            (6,	42.10)
            (7,	42.07)
            (8,	42.03)
            };
            \addlegendentry{$P_{drop}=.1$}
            \addplot[color=clr4, line width=.3mm, mark=square] coordinates {
            (1,	41.73)
            (2,	41.93)
            (3,	42.10)
            (4,	42.07)
            (5,	42.07)
            (6,	42.10)
            (7,	42.03)
            (8,	42.00)
            };
            \addlegendentry{$P_{drop}=.2$}
            \addplot[color=clr5, line width=.3mm, mark=square] coordinates {
            (1,	41.47)
            (2,	41.80)
            (3,	41.83)
            (4,	41.87)
            (5,	41.90)
            (6,	41.93)
            (7,	41.93)
            (8,	41.97)
            };
            \addlegendentry{$P_{drop}=.3$}
            \end{axis}
            \end{tikzpicture}
        }
    \end{subfigure}
    \caption{Beam width vs. SacreBLEU on WMT 2014 benchmarks using Word Drop Regularization.
    Reported scores are averages over n=3 independent training runs.
    See Appendix Tables~\ref{worddropfr} and~\ref{worddropde} for results with standard error measurements.}
    \label{worddrop}
\end{figure*}

\subsubsection{Top-K Approximation}
Up to this point, TeaForN setups have used Eq.~\eqref{eq:teaforn} to approximate the inference-time decoder input.

We now share results for an alternative approximation called Top-K, described by Eq.~\eqref{eq:e2e_k} and inspired by~\cite{mixer15}, which feeds the embedding expectation of the decoder output. If $K=V$, Top-K is an exact expectation.
If $K<V$, Top-K approximates the expectation as the probability-weighted embeddings of the $K$ most likely outputs.

In this experiment, we try $K \in \{4,V\}$ and $N \in \{2, 3\}$ using  Transformer\textsubscript{big} as our base model.
We report results on both WMT14 benchmarks.
We use discount factor $\lambda=.2$ for all setups.

Figure~\ref{topkplot} shows that Top-K does not work as well as the original TeaForN approximation described by Eq.~\eqref{eq:teaforn}.
Top-K with $K=4$ performs worse than TeaForN on the En-De benchmark but not the En-Fr benchmark.
When $K=V$, the situation is the exact opposite, with Top-K performing better on the En-De benchmark but not the En-Fr benchmark.

\subsubsection{Word Drop Regularization}

TeaForN could potentially have regularization-like effects by solving for a more difficult task than standard teacher-forcing.
TeaForN trains models to decode not just from groundtruth prefixes, but also from past model predictions.

To see whether regularization-like effects are responsible for the gains seen using TeaForN, we perform a regularization experiment using Transformer\textsubscript{big}.
In particular, we randomly sample a set of groundtruth decoder input words in each example with probability $P_{drop} \in \{0,.1,.2,.3\}$. For each selected word, we apply the $word\_drop$ regularization by masking all its embedding elements to zero.

Fig.~\ref{worddrop} shows that word drop regularization increases performance against the En-De benchmark but reduces performance against En-Fr.
These results are in stark contrast with the results of TeaForN, which only improves performance in the En-Fr case.
Though TeaForN may have regularization-like effects, they are likely different from the effects of word drop regularization.

\subsubsection{Additional Compute}

TeaForN uses more compute resources than Teacher-forcing when inference-time architecture and number of training iterations are the same, as is the case in our Transformer\textsubscript{big} experiments.

To enable a fair comparison in terms of training-time compute, we conduct an experiment where we train Transformer\textsubscript{big} so that the total device time is about the same.
We train the baseline for 1.5x iterations, a figure which was estimated from the observed training speeds of Transformer\textsubscript{big} and TeaFor2 (4.5 iterations/sec and 6.8 iterations/sec).

Table~\ref{samecompute} shows that this additional training does not significantly benefit Transformer\textsubscript{big}, for either language pair.
Based on these results, we conclude that the benefits of TeaForN do not likely derive from additional compute.

\subsection{News Summarization}
We now present our experiment on News Summarization using PEGASUS\textsubscript{large}~\cite{Zhang2019} as our base model.

We test on two News Summarization tasks, CNN/Dailymail and Gigaword:
\begin{itemize}
    \item \textbf{CNN/Dailymail}~\cite{Hermann2015} consists of 93k CNN articles and 220k Daily Mail articles, where publishers provide bullet-style summaries with each article.
    \item \textbf{Gigaword}~\cite{Rush2015} contains 4M articles from seven publishers, where article headlines serve as the summary.
\end{itemize}

\begin{table*}[!hpbt]
\begin{centering}
\begin{tabular}{ @{} rcc @{} } \toprule
\ra{1.3}
R1/R2/RL & CNN/Dailymail & Gigaword \\
\midrule
BERTShare~\cite{Rothe2019}        & 39.25/18.09/36.45 & 38.13/19.81/35.62 \\
MASS~\cite{Song2019}             & 42.12/19.50/39.01 & 38.73/19.71/35.96 \\
UniLM~\cite{Dong2019}            & 43.33/20.21/40.51 & 38.45/19.45/35.75 \\
BART~\cite{Lewis2019}             & 44.16/21.28/40.90 & - \\
T5~\cite{Raffel2019}               & 43.52/21.55/40.69 & - \\
PEGASUS~\cite{Zhang2019}          & 44.17/21.47/41.11 & 39.12/19.86/36.24 \\
\hline
 (Greedy) TeaFor3+PEGASUS              & 43.90/20.36/41.20 & 39.10/19.40/36.30 \\
(Beam@k=8) TeaFor3+PEGASUS            & 44.20/\textbf{21.70}/\textbf{41.32} & 39.16/\textbf{20.16}/\textbf{36.54} \\
 \bottomrule
\end{tabular}
\caption{
A comparison of News Summarization models on CNN/Dailymail and Gigaword benchmarks.
Scores are ROUGE-1/ROUGE-2/ROUGE-L F-measures.
PEGASUS is shorthand for PEGASUS\textsubscript{large} (HugeNews) and uses beam width $k=8$ for both tasks.
We use TeaFor3 with $\lambda=.5$ and weight-sharing.
}
\label{sumleaderboard}
\end{centering}
\end{table*}

\begin{table*}[!htpb]
\begin{centering}
\begin{tabular}{ @{} r c c c @{} } \toprule
\ra{1.3}
                & Decoder Layers & Steps/sec & HBM usage (GB) \\
\hline
Teacher-Forcing & 16             & 1.97          &  8.64  \\ 
TeaFor2         & 32             & 1.25          &  8.96  \\ 
TeaFor3         & 48             & .986          &  10.33 \\
\bottomrule
\end{tabular}
\caption{
Performance of TeaForN with weight-sharing during PEGASUS\textsubscript{large} (HugeNews) pretraining.
Steps/sec refers to the number of training batches processed per second.
High-Bandwidth Memory usage refers to the consumption of Google Cloud TPU device memory.
}
\label{perftable}
\end{centering}
\end{table*}

The PEGASUS approach has been shown to work better on News Summarization tasks when pretrained on HugeNews, a dataset of 1.5B news-like articles scraped from the web between 2013 and 2019.
We use the same pretraining procedure as originally described for PEGASUS\textsubscript{large} (HugeNews), which uses teacher-forcing to learn based on an unsupervised Gap Sentence Generation task~\cite{Zhang2019}.

We use TeaFor3 with $\lambda=.5$ and weight-sharing.
For model selection, we use the checkpoint with the highest ROUGE-L F-score on the validation set, with evaluations every 1k steps.
We stop training on Gigaword after 160k steps and CNN/Dailymail after 400k steps.

Final scores in Table~\ref{sumleaderboard} show the benefits of TeaForN on summarization tasks.
Using just greedy decoding, TeaFor3 setups match or exceed the previous state-of-the-art ROUGE-L score on both CNN/Dailymail and Gigaword benchmarks, with an 8x cheaper decoder.
Using beam search, TeaForN increases performance on the CNN/Dailymail task by +.23 ROUGE-2 (21.70 vs 21.47) and +.21 ROUGE-L (41.32 vs 41.11) and the Gigaword task by +.30 ROUGE-2 (20.16 vs 19.86) and +.30 ROUGE-L (36.54 vs 36.24).

\subsection{Training Performance}
Table~\ref{perftable} shows how TeaForN affects training performance, using Google Cloud TPUs~\cite{YouZHDK19}.
TeaFor2 slows down training by 37\% compared to standard teacher-forcing (1.25 steps/sec vs 1.97) and TeaFor3 by 50\% (.986 steps/sec vs 1.97).
TeaFor2 increases High-Bandwidth Memory (HBM) usage by 4\% compared to teacher-forcing (8.96GB vs 8.64) and TeaFor3 by 20\%.

While training cost and speed are moderately impacted by TeaForN, we note that inference cost is significantly reduced by virtue of producing models that reach similar quality with fewer beams, enabling significant cost savings for production models, in addition to overall stronger models.

\section{Conclusion}
In this work, we introduce a new technique for sequence generation models called Teacher-Forcing with N-grams (TeaForN), which
(a) addresses exposure bias,
(b) allows the decoder to better take into account future decisions, and
(c) requires no curriculum training.

We show empirical evidence of the efficacy of TeaForN on several sequence generation tasks.
With Transformer~\textsubscript{big}~\cite{vaswani2017attention}, we boost the performance of Transformer\textsubscript{big} significantly on the En-Fr benchmark.
With PEGASUS~\textsubscript{large}~\cite{Zhang2019}, we improve upon the existing ROUGE-L scores the Gigaword and CNN/Dailymail benchmarks~\cite{Rush2015}.
Further, we show that TeaForN can match the prior state-of-the-art ROUGE-L scores on the summarization benchmarks without beam search, representing an 8x reduction in decoder cost at inference.

Overall, TeaForN is a promising approach for improving quality and/or reducing inference costs in sequence generation models.

\bibliography{emnlp2020}
\bibliographystyle{acl_natbib}

\appendix

\begin{table*}[!hpbt]
\begin{centering}
\begin{tabular}{ @{} rcccccc @{} } \toprule
\ra{1.3}
$k$ & $\lambda=0$ & $\lambda=0.2$ & $\lambda=0.4$ & $\lambda=0.6$ & $\lambda=0.8$ & $\lambda=1$ \\
\midrule
1 & $41.83 \pm 0.05$ & $42.17 \pm 0.07$ & $42.10 \pm 0.08$ & $42.07 \pm 0.07$ & $42.00 \pm 0.05$ & $42.03 \pm 0.05$ \\
2 & $42.20 \pm 0.09$ & $42.60 \pm 0.05$ & $42.50 \pm 0.08$ & $42.30 \pm 0.05$ & $42.33 \pm 0.10$ & $42.43 \pm 0.07$ \\
3 & $42.37 \pm 0.07$ & $42.67 \pm 0.03$ & $42.63 \pm 0.05$ & $42.43 \pm 0.03$ & $42.53 \pm 0.05$ & $42.57 \pm 0.03$ \\
4 & $42.40 \pm 0.05$ & $42.73 \pm 0.05$ & $42.63 \pm 0.03$ & $42.47 \pm 0.03$ & $42.63 \pm 0.03$ & $42.60 \pm 0.05$ \\
5 & $42.30 \pm 0.08$ & $42.80 \pm 0.05$ & $42.70 \pm 0.05$ & $42.47 \pm 0.03$ & $42.57 \pm 0.03$ & $42.57 \pm 0.03$ \\
6 & $42.37 \pm 0.07$ & $42.77 \pm 0.03$ & $42.63 \pm 0.03$ & $42.40 \pm 0.00$ & $42.60 \pm 0.00$ & $42.60 \pm 0.05$ \\
7 & $42.33 \pm 0.07$ & $42.73 \pm 0.05$ & $42.63 \pm 0.03$ & $42.40 \pm 0.00$ & $42.57 \pm 0.03$ & $42.50 \pm 0.05$ \\
8 & $42.33 \pm 0.07$ & $\bm{42.73 \pm 0.05}$ & $42.60 \pm 0.05$ & $42.43 \pm 0.03$ & $42.50 \pm 0.05$ & $42.50 \pm 0.05$ \\
\bottomrule
\end{tabular}
\caption{
Mean SacreBLEU and Standard Error (n=3) of TeaFor2 on the WMT14 English-French benchmark, for all beam sizes $k$ and discount factors $\lambda$ tested.
Bold font indicates the configuration reported in Table~\ref{translateleaderboard}.
}
\label{teafor2enfr14}
\end{centering}
\end{table*}

\begin{table*}[!hpbt]
\begin{centering}
\begin{tabular}{ @{} rcccccc @{} } \toprule
\ra{1.3}
$k$ & $\lambda=0$ & $\lambda=0.2$ & $\lambda=0.4$ & $\lambda=0.6$ & $\lambda=0.8$ & $\lambda=1$ \\
\midrule
1 & $31.10 \pm 0.00$ & $31.67 \pm 0.03$ & $31.53 \pm 0.03$ & $31.53 \pm 0.03$ & $31.63 \pm 0.14$ & $31.57 \pm 0.05$ \\
2 & $31.40 \pm 0.05$ & $32.00 \pm 0.05$ & $31.83 \pm 0.07$ & $31.97 \pm 0.03$ & $31.97 \pm 0.11$ & $31.97 \pm 0.05$ \\
3 & $31.50 \pm 0.05$ & $32.10 \pm 0.05$ & $31.97 \pm 0.05$ & $31.90 \pm 0.05$ & $32.00 \pm 0.12$ & $32.03 \pm 0.03$ \\
4 & $31.53 \pm 0.03$ & $32.10 \pm 0.05$ & $31.90 \pm 0.05$ & $31.97 \pm 0.05$ & $32.03 \pm 0.10$ & $32.03 \pm 0.03$ \\
5 & $31.57 \pm 0.03$ & $32.07 \pm 0.03$ & $31.93 \pm 0.03$ & $31.97 \pm 0.05$ & $32.03 \pm 0.10$ & $32.00 \pm 0.00$ \\
6 & $31.47 \pm 0.05$ & $32.10 \pm 0.08$ & $31.90 \pm 0.05$ & $31.97 \pm 0.03$ & $32.00 \pm 0.08$ & $31.93 \pm 0.03$ \\
7 & $31.43 \pm 0.07$ & $32.07 \pm 0.07$ & $31.87 \pm 0.03$ & $31.87 \pm 0.03$ & $31.97 \pm 0.11$ & $31.93 \pm 0.03$ \\
8 & $31.43 \pm 0.03$ & $\bm{32.07 \pm 0.07}$ & $31.87 \pm 0.10$ & $31.87 \pm 0.07$ & $31.93 \pm 0.14$ & $31.90 \pm 0.00$ \\
\bottomrule
\end{tabular}
\caption{
Mean SacreBLEU and Standard Error (n=3) of TeaFor2 on the WMT12 English-French benchmark, for all beam sizes $k$ and discount factors $\lambda$ tested.
Bold font indicates the configuration reported in Table~\ref{translateleaderboard}.
}
\label{teafor2enfr12}
\end{centering}
\end{table*}

\begin{table*}[!hpbt]
\begin{centering}
\begin{tabular}{ @{} rcccccc @{} } \toprule
\ra{1.3}
$k$ & $\lambda=0$ & $\lambda=0.2$ & $\lambda=0.4$ & $\lambda=0.6$ & $\lambda=0.8$ & $\lambda=1$ \\
\midrule
1 & $41.83 \pm 0.05$ & $42.07 \pm 0.05$ & $41.97 \pm 0.03$ & $42.17 \pm 0.07$ & $42.00 \pm 0.05$ & $42.00 \pm 0.05$ \\
2 & $42.20 \pm 0.09$ & $42.37 \pm 0.10$ & $42.27 \pm 0.05$ & $42.40 \pm 0.08$ & $42.30 \pm 0.05$ & $42.30 \pm 0.08$ \\
3 & $42.37 \pm 0.07$ & $42.53 \pm 0.03$ & $42.37 \pm 0.03$ & $42.53 \pm 0.05$ & $42.43 \pm 0.05$ & $42.40 \pm 0.08$ \\
4 & $42.40 \pm 0.05$ & $42.63 \pm 0.07$ & $42.37 \pm 0.03$ & $42.57 \pm 0.03$ & $42.43 \pm 0.10$ & $42.43 \pm 0.03$ \\
5 & $42.30 \pm 0.08$ & $42.57 \pm 0.12$ & $42.43 \pm 0.03$ & $42.57 \pm 0.03$ & $42.47 \pm 0.12$ & $42.50 \pm 0.05$ \\
6 & $42.37 \pm 0.07$ & $42.60 \pm 0.08$ & $42.43 \pm 0.03$ & $42.50 \pm 0.05$ & $42.50 \pm 0.05$ & $42.43 \pm 0.03$ \\
7 & $42.33 \pm 0.07$ & $42.60 \pm 0.09$ & $42.47 \pm 0.03$ & $42.50 \pm 0.05$ & $42.47 \pm 0.07$ & $42.47 \pm 0.03$ \\
8 & $42.33 \pm 0.07$ & $42.57 \pm 0.10$ & $\bm{42.43 \pm 0.03}$ & $42.47 \pm 0.03$ & $42.40 \pm 0.05$ & $42.47 \pm 0.03$ \\
\bottomrule
\end{tabular}
\caption{
Mean SacreBLEU and Standard Error (n=3) of TeaFor3 on the WMT14 English-French benchmark, for all beam sizes $k$ and discount factors $\lambda$ tested.
Bold font indicates the configuration reported in Table~\ref{translateleaderboard}.
}
\label{teafor3enfr14}
\end{centering}
\end{table*}

\begin{table*}[!hpbt]
\begin{centering}
\begin{tabular}{ @{} rcccccc @{} } \toprule
\ra{1.3}
$k$ & $\lambda=0$ & $\lambda=0.2$ & $\lambda=0.4$ & $\lambda=0.6$ & $\lambda=0.8$ & $\lambda=1$ \\
\midrule
1 & $31.10 \pm 0.00$ & $31.40 \pm 0.00$ & $31.70 \pm 0.05$ & $31.60 \pm 0.05$ & $31.50 \pm 0.08$ & $31.57 \pm 0.05$ \\
2 & $31.40 \pm 0.05$ & $31.80 \pm 0.05$ & $32.00 \pm 0.05$ & $31.83 \pm 0.07$ & $31.80 \pm 0.08$ & $31.83 \pm 0.03$ \\
3 & $31.50 \pm 0.05$ & $31.83 \pm 0.05$ & $32.10 \pm 0.05$ & $31.93 \pm 0.10$ & $31.80 \pm 0.09$ & $32.00 \pm 0.05$ \\
4 & $31.53 \pm 0.03$ & $31.87 \pm 0.03$ & $32.03 \pm 0.07$ & $31.93 \pm 0.10$ & $31.80 \pm 0.09$ & $31.97 \pm 0.12$ \\
5 & $31.57 \pm 0.03$ & $31.87 \pm 0.03$ & $32.00 \pm 0.09$ & $31.90 \pm 0.09$ & $31.80 \pm 0.14$ & $31.93 \pm 0.07$ \\
6 & $31.47 \pm 0.05$ & $31.83 \pm 0.03$ & $31.93 \pm 0.05$ & $31.87 \pm 0.10$ & $31.77 \pm 0.10$ & $31.93 \pm 0.10$ \\
7 & $31.43 \pm 0.07$ & $31.80 \pm 0.00$ & $31.93 \pm 0.03$ & $31.87 \pm 0.10$ & $31.80 \pm 0.09$ & $31.93 \pm 0.10$ \\
8 & $31.43 \pm 0.03$ & $31.77 \pm 0.03$ & $\bm{31.90 \pm 0.05}$ & $31.87 \pm 0.10$ & $31.77 \pm 0.10$ & $31.83 \pm 0.07$ \\
\bottomrule
\end{tabular}
\caption{
Mean SacreBLEU and Standard Error (n=3) of TeaFor3 on the WMT12 English-French benchmark, for all beam sizes $k$ and discount factors $\lambda$ tested.
Bold font indicates the configuration reported in Table~\ref{translateleaderboard}.
}
\label{teafor3enfr12}
\end{centering}
\end{table*}

\begin{table*}[!hpbt]
\begin{centering}
\begin{tabular}{ @{} rcccccc @{} } \toprule
\ra{1.3}
$k$ & $\lambda=0$ & $\lambda=0.2$ & $\lambda=0.4$ & $\lambda=0.6$ & $\lambda=0.8$ & $\lambda=1$ \\
\midrule
1 & $28.10 \pm 0.08$ & $28.17 \pm 0.07$ & $28.17 \pm 0.03$ & $28.27 \pm 0.10$ & $28.10 \pm 0.08$ & $27.83 \pm 0.07$ \\
2 & $28.67 \pm 0.07$ & $28.67 \pm 0.07$ & $28.80 \pm 0.08$ & $28.87 \pm 0.07$ & $28.83 \pm 0.03$ & $28.47 \pm 0.07$ \\
3 & $28.97 \pm 0.10$ & $28.97 \pm 0.07$ & $29.03 \pm 0.05$ & $29.07 \pm 0.03$ & $29.00 \pm 0.08$ & $28.73 \pm 0.07$ \\
4 & $29.07 \pm 0.10$ & $29.07 \pm 0.07$ & $29.27 \pm 0.07$ & $29.13 \pm 0.03$ & $29.13 \pm 0.07$ & $28.77 \pm 0.07$ \\
5 & $29.20 \pm 0.12$ & $29.13 \pm 0.10$ & $29.23 \pm 0.03$ & $29.17 \pm 0.05$ & $29.20 \pm 0.08$ & $28.80 \pm 0.05$ \\
6 & $29.30 \pm 0.12$ & $29.17 \pm 0.07$ & $29.27 \pm 0.03$ & $29.20 \pm 0.00$ & $29.23 \pm 0.07$ & $28.80 \pm 0.05$ \\
7 & $29.23 \pm 0.12$ & $29.20 \pm 0.05$ & $29.30 \pm 0.05$ & $29.20 \pm 0.05$ & $29.27 \pm 0.05$ & $28.87 \pm 0.03$ \\
8 & $29.20 \pm 0.12$ & $29.20 \pm 0.05$ & $\bm{29.30 \pm 0.05}$ & $29.13 \pm 0.03$ & $29.23 \pm 0.07$ & $28.80 \pm 0.00$ \\
\bottomrule
\end{tabular}
\caption{
Mean SacreBLEU and Standard Error (n=3) of TeaFor2 on the WMT14 English-German benchmark, for all beam sizes $k$ and discount factors $\lambda$ tested.
Bold font indicates the configuration reported in Table~\ref{translateleaderboard}.
}
\label{teafor2ende14}
\end{centering}
\end{table*}

\begin{table*}[!hpbt]
\begin{centering}
\begin{tabular}{ @{} rcccccc @{} } \toprule
\ra{1.3}
$k$ & $\lambda=0$ & $\lambda=0.2$ & $\lambda=0.4$ & $\lambda=0.6$ & $\lambda=0.8$ & $\lambda=1$ \\
\midrule
1 & $22.07 \pm 0.03$ & $22.10 \pm 0.05$ & $22.20 \pm 0.09$ & $22.10 \pm 0.05$ & $22.13 \pm 0.03$ & $21.97 \pm 0.07$ \\
2 & $22.40 \pm 0.05$ & $22.50 \pm 0.12$ & $22.50 \pm 0.05$ & $22.30 \pm 0.00$ & $22.37 \pm 0.03$ & $22.40 \pm 0.05$ \\
3 & $22.50 \pm 0.05$ & $22.57 \pm 0.07$ & $22.60 \pm 0.08$ & $22.30 \pm 0.05$ & $22.47 \pm 0.03$ & $22.43 \pm 0.07$ \\
4 & $22.43 \pm 0.07$ & $22.57 \pm 0.07$ & $22.57 \pm 0.10$ & $22.30 \pm 0.05$ & $22.37 \pm 0.03$ & $22.43 \pm 0.07$ \\
5 & $22.50 \pm 0.09$ & $22.57 \pm 0.07$ & $22.53 \pm 0.07$ & $22.37 \pm 0.05$ & $22.37 \pm 0.03$ & $22.37 \pm 0.05$ \\
6 & $22.43 \pm 0.07$ & $22.57 \pm 0.07$ & $22.53 \pm 0.07$ & $22.33 \pm 0.07$ & $22.30 \pm 0.05$ & $22.40 \pm 0.05$ \\
7 & $22.43 \pm 0.07$ & $22.53 \pm 0.10$ & $22.47 \pm 0.10$ & $22.30 \pm 0.05$ & $22.30 \pm 0.05$ & $22.37 \pm 0.05$ \\
8 & $22.37 \pm 0.10$ & $22.43 \pm 0.10$ & $\bm{22.53 \pm 0.07}$ & $22.30 \pm 0.05$ & $22.27 \pm 0.03$ & $22.30 \pm 0.05$ \\
\bottomrule
\end{tabular}
\caption{
Mean SacreBLEU and Standard Error (n=3) of TeaFor2 on the WMT12 English-German benchmark, for all beam sizes $k$ and discount factors $\lambda$ tested.
Bold font indicates the configuration reported in Table~\ref{translateleaderboard}.
}
\label{teafor2ende12}
\end{centering}
\end{table*}

\begin{table*}[!hpbt]
\begin{centering}
\begin{tabular}{ @{} rcccccc @{} } \toprule
\ra{1.3}
$k$ & $\lambda=0$ & $\lambda=0.2$ & $\lambda=0.4$ & $\lambda=0.6$ & $\lambda=0.8$ & $\lambda=1$ \\
\midrule
1 & $28.10 \pm 0.08$ & $28.40 \pm 0.12$ & $28.07 \pm 0.14$ & $28.23 \pm 0.05$ & $28.17 \pm 0.10$ & $27.80 \pm 0.05$ \\
2 & $28.67 \pm 0.07$ & $29.00 \pm 0.08$ & $28.67 \pm 0.07$ & $28.73 \pm 0.03$ & $28.60 \pm 0.08$ & $28.40 \pm 0.05$ \\
3 & $28.97 \pm 0.10$ & $29.27 \pm 0.05$ & $28.87 \pm 0.07$ & $29.00 \pm 0.00$ & $28.87 \pm 0.05$ & $28.63 \pm 0.03$ \\
4 & $29.07 \pm 0.10$ & $29.33 \pm 0.07$ & $28.97 \pm 0.07$ & $29.00 \pm 0.05$ & $28.93 \pm 0.07$ & $28.67 \pm 0.05$ \\
5 & $29.20 \pm 0.12$ & $29.30 \pm 0.09$ & $29.13 \pm 0.05$ & $29.00 \pm 0.00$ & $28.97 \pm 0.07$ & $28.70 \pm 0.05$ \\
6 & $29.30 \pm 0.12$ & $29.33 \pm 0.07$ & $29.17 \pm 0.03$ & $29.03 \pm 0.03$ & $29.00 \pm 0.05$ & $28.77 \pm 0.10$ \\
7 & $29.23 \pm 0.12$ & $29.37 \pm 0.07$ & $29.20 \pm 0.08$ & $29.10 \pm 0.05$ & $29.03 \pm 0.05$ & $28.73 \pm 0.07$ \\
8 & $29.20 \pm 0.12$ & $29.33 \pm 0.10$ & $\bm{29.23 \pm 0.05}$ & $29.07 \pm 0.05$ & $29.00 \pm 0.05$ & $28.77 \pm 0.10$ \\
\bottomrule
\end{tabular}
\caption{
Mean SacreBLEU and Standard Error (n=3) of TeaFor3 on the WMT14 English-German benchmark, for all beam sizes $k$ and discount factors $\lambda$ tested.
Bold font indicates the configuration reported in Table~\ref{translateleaderboard}.
}
\label{teafor3ende14}
\end{centering}
\end{table*}

\begin{table*}[!hpbt]
\begin{centering}
\begin{tabular}{ @{} rcccccc @{} } \toprule
\ra{1.3}
$k$ & $\lambda=0$ & $\lambda=0.2$ & $\lambda=0.4$ & $\lambda=0.6$ & $\lambda=0.8$ & $\lambda=1$ \\
\midrule
1 & $22.07 \pm 0.03$ & $22.17 \pm 0.03$ & $22.07 \pm 0.03$ & $22.10 \pm 0.05$ & $22.07 \pm 0.03$ & $22.03 \pm 0.03$ \\
2 & $22.40 \pm 0.05$ & $22.43 \pm 0.05$ & $22.50 \pm 0.05$ & $22.40 \pm 0.00$ & $22.43 \pm 0.03$ & $22.30 \pm 0.05$ \\
3 & $22.50 \pm 0.05$ & $22.33 \pm 0.03$ & $22.50 \pm 0.05$ & $22.43 \pm 0.03$ & $22.40 \pm 0.00$ & $22.37 \pm 0.03$ \\
4 & $22.43 \pm 0.07$ & $22.37 \pm 0.03$ & $22.57 \pm 0.05$ & $22.43 \pm 0.05$ & $22.33 \pm 0.03$ & $22.27 \pm 0.03$ \\
5 & $22.50 \pm 0.09$ & $22.30 \pm 0.00$ & $22.50 \pm 0.09$ & $22.47 \pm 0.07$ & $22.33 \pm 0.03$ & $22.27 \pm 0.05$ \\
6 & $22.43 \pm 0.07$ & $22.27 \pm 0.03$ & $22.47 \pm 0.10$ & $22.37 \pm 0.05$ & $22.30 \pm 0.00$ & $22.20 \pm 0.05$ \\
7 & $22.43 \pm 0.07$ & $22.20 \pm 0.00$ & $22.43 \pm 0.07$ & $22.40 \pm 0.05$ & $22.30 \pm 0.00$ & $22.17 \pm 0.03$ \\
8 & $22.37 \pm 0.10$ & $22.20 \pm 0.00$ & $\bm{22.40 \pm 0.05}$ & $22.33 \pm 0.05$ & $22.20 \pm 0.00$ & $22.13 \pm 0.05$ \\
\bottomrule
\end{tabular}
\caption{
Mean SacreBLEU and Standard Error (n=3) of TeaFor3 on the WMT12 English-German benchmark, for all beam sizes $k$ and discount factors $\lambda$ tested.
Bold font indicates the configuration reported in Table~\ref{translateleaderboard}.
}
\label{teafor3ende12}
\end{centering}
\end{table*}

\begin{table*}[!hpbt]
\begin{centering}
\begin{tabular}{ @{} rcccccc @{} } \toprule
\ra{1.3}
$k$ & Teacher-forcing & TeaForN & Top-4 & Top-V \\
\midrule
1 & $41.83 \pm 0.05$ & $42.17 \pm 0.07$ & $42.23 \pm 0.11$ & $41.90 \pm 0.08$ \\
2 & $42.20 \pm 0.09$ & $42.60 \pm 0.05$ & $42.50 \pm 0.16$ & $42.27 \pm 0.05$ \\
3 & $42.37 \pm 0.07$ & $42.67 \pm 0.03$ & $42.73 \pm 0.12$ & $42.47 \pm 0.05$ \\
4 & $42.40 \pm 0.05$ & $42.73 \pm 0.05$ & $42.77 \pm 0.14$ & $42.47 \pm 0.05$ \\
5 & $42.30 \pm 0.08$ & $42.80 \pm 0.05$ & $42.77 \pm 0.10$ & $42.47 \pm 0.05$ \\
6 & $42.37 \pm 0.07$ & $42.77 \pm 0.03$ & $42.80 \pm 0.12$ & $42.47 \pm 0.05$ \\
7 & $42.33 \pm 0.07$ & $42.73 \pm 0.05$ & $42.73 \pm 0.12$ & $42.47 \pm 0.07$ \\
8 & $42.33 \pm 0.07$ & $42.73 \pm 0.05$ & $42.73 \pm 0.12$ & $42.43 \pm 0.07$ \\
\bottomrule
\end{tabular}
\caption{
Mean SacreBLEU and Standard Error (n=3) of TeaFor2 ($\lambda=.2$) on the WMT14 English-French benchmark using different approximation methods, for all beam sizes $k$.
}
\label{approxenfr}
\end{centering}
\end{table*}

\begin{table*}[!hpbt]
\begin{centering}
\begin{tabular}{ @{} rcccccc @{} } \toprule
\ra{1.3}
$k$ & Teacher-forcing & TeaForN & Top-4 & Top-V \\
\midrule
1 & $28.10 \pm 0.08$ & $28.17 \pm 0.07$ & $27.83 \pm 0.10$ & $28.03 \pm 0.10$ \\
2 & $28.67 \pm 0.07$ & $28.67 \pm 0.07$ & $28.57 \pm 0.10$ & $28.73 \pm 0.07$ \\
3 & $28.97 \pm 0.10$ & $28.97 \pm 0.07$ & $28.80 \pm 0.05$ & $29.00 \pm 0.00$ \\
4 & $29.07 \pm 0.10$ & $29.07 \pm 0.07$ & $28.90 \pm 0.05$ & $29.13 \pm 0.03$ \\
5 & $29.20 \pm 0.12$ & $29.13 \pm 0.10$ & $28.97 \pm 0.05$ & $29.23 \pm 0.03$ \\
6 & $29.30 \pm 0.12$ & $29.17 \pm 0.07$ & $29.00 \pm 0.08$ & $29.30 \pm 0.05$ \\
7 & $29.23 \pm 0.12$ & $29.20 \pm 0.05$ & $29.00 \pm 0.08$ & $29.23 \pm 0.07$ \\
8 & $29.20 \pm 0.12$ & $29.20 \pm 0.05$ & $29.00 \pm 0.05$ & $29.30 \pm 0.05$ \\
\bottomrule
\end{tabular}
\caption{
Mean SacreBLEU and Standard Error (n=3) of TeaFor2 ($\lambda=.2$) on the WMT14 English-German benchmark using different approximation methods, for all beam sizes $k$.
}
\label{approxende}
\end{centering}
\end{table*}

\begin{table*}[!hpbt]
\begin{centering}
\begin{tabular}{ @{} rcccc @{} } \toprule
\ra{1.3}
$k$ & Transformer\textsubscript{big} & $P_{drop}=.01$ & $P_{drop}=.02$ & $P_{drop}=.03$ \\
\midrule
1 & $41.83 \pm 0.05$ & $41.63 \pm 0.07$ & $41.73 \pm 0.12$ & $41.47 \pm 0.03$ \\
2 & $42.20 \pm 0.09$ & $41.93 \pm 0.07$ & $41.93 \pm 0.07$ & $41.80 \pm 0.08$ \\
3 & $42.37 \pm 0.07$ & $42.13 \pm 0.05$ & $42.10 \pm 0.09$ & $41.83 \pm 0.07$ \\
4 & $42.40 \pm 0.05$ & $42.17 \pm 0.07$ & $42.07 \pm 0.12$ & $41.87 \pm 0.07$ \\
5 & $42.30 \pm 0.08$ & $42.10 \pm 0.05$ & $42.07 \pm 0.07$ & $41.90 \pm 0.09$ \\
6 & $42.37 \pm 0.07$ & $42.10 \pm 0.05$ & $42.10 \pm 0.08$ & $41.93 \pm 0.07$ \\
7 & $42.33 \pm 0.07$ & $42.07 \pm 0.07$ & $42.03 \pm 0.05$ & $41.97 \pm 0.10$ \\
8 & $42.33 \pm 0.07$ & $42.03 \pm 0.07$ & $42.00 \pm 0.08$ & $41.97 \pm 0.12$ \\
\bottomrule
\end{tabular}
\caption{
Mean SacreBLEU and Standard Error (n=3) on the WMT14 English-French benchmark using word drop regularization, for all beam sizes $k$.
}
\label{worddropfr}
\end{centering}
\end{table*}

\begin{table*}[!hpbt]
\begin{centering}
\begin{tabular}{ @{} rcccccc @{} } \toprule
\ra{1.3}& Transformer\textsubscript{big} & $P_{drop}=.01$ & $P_{drop}=.02$ & $P_{drop}=.03$ \\
\midrule
1 & $28.10 \pm 0.08$ & $28.60 \pm 0.12$ & $28.37 \pm 0.07$ & $28.20 \pm 0.08$ \\
2 & $28.67 \pm 0.07$ & $29.13 \pm 0.15$ & $28.90 \pm 0.00$ & $28.77 \pm 0.07$ \\
3 & $28.97 \pm 0.10$ & $29.23 \pm 0.20$ & $29.13 \pm 0.03$ & $28.93 \pm 0.05$ \\
4 & $29.07 \pm 0.10$ & $29.43 \pm 0.17$ & $29.20 \pm 0.05$ & $29.07 \pm 0.03$ \\
5 & $29.20 \pm 0.12$ & $29.47 \pm 0.10$ & $29.33 \pm 0.03$ & $29.13 \pm 0.03$ \\
6 & $29.30 \pm 0.12$ & $29.50 \pm 0.12$ & $29.30 \pm 0.00$ & $29.23 \pm 0.03$ \\
7 & $29.23 \pm 0.12$ & $29.47 \pm 0.12$ & $29.30 \pm 0.00$ & $29.20 \pm 0.05$ \\
8 & $29.20 \pm 0.12$ & $29.47 \pm 0.10$ & $29.33 \pm 0.03$ & $29.20 \pm 0.05$ \\

\bottomrule
\end{tabular}
\caption{
Mean SacreBLEU and Standard Error (n=3) on the WMT14 English-German benchmark using word drop regularization, for all beam sizes $k$.
}
\label{worddropde}
\end{centering}
\end{table*}

\end{document}